\pdfoutput=1

\documentclass[11pt]{article}

\usepackage{latex/acl}

\usepackage{times}
\usepackage{latexsym}

\usepackage[T1]{fontenc}

\usepackage[utf8]{inputenc}

\usepackage{microtype}

\usepackage{inconsolata}
\usepackage{graphicx}
\graphicspath{ {./images/} }

\usepackage{threeparttable}

\title{ALMs: Authorial Language Models for Authorship Attribution }

\author{Weihang Huang \and Akira Murakami \and Jack Grieve  \\
  Department of English Language and Linguistics, University of Birmingham \\
  \texttt{wxh207@student.bham.ac.uk}  
  \texttt{a.murakami@bham.ac.uk}
  \texttt{j.grieve@bham.ac.uk}}

\begin{document}
\maketitle
\begin{abstract}
In this paper, we introduce an authorship attribution method called Authorial Language Models (ALMs) that involves identifying the most likely author of a questioned document based on the perplexity of the questioned document calculated for a set of causal language models fine-tuned on the writings of a set of candidate author. We benchmarked ALMs against state-of-art-systems using the CCAT50 dataset and the Blogs50 datasets. We find that ALMs achieves a macro-average accuracy score of 83.6\% on Blogs50, outperforming all other methods, and 74.9\% on CCAT50, matching the performance of the best method. To assess the performance of ALMs on shorter texts, we also conducted text ablation testing. We found that to reach a macro-average accuracy of 70\%, ALMs needs 40 tokens on Blogs50 and 400 tokens on CCAT50, while to reach 60\% ALMs requires 20 tokens on Blogs50 and 70 tokens on CCAT50.

\end{abstract}

\section{Introduction}

For over a century, researchers have developed methods for \textit{authorship attribution} to resolve cases of disputed authorship by comparing the style of a questioned document to writing samples from a set of candidate authors \citep{Juola_2006,Stamatatos_2009}. The goal of authorship attribution is to identify the candidate whose style of writing is most similar to a questioned document. Stylometry is the quantitative analysis of style and is a common approach to authorship attribution \citep{Juola_2006,Stamatatos_2009}. A wide range of different measurements and methods for authorship attribution have been developed in stylometry \citep{Grieve_2007,Stamatatos_2009}. Popular techniques include Principal Component Analysis of function word frequencies \citep{Binongo_2003,grieve2023_language} and distance-based comparisons of the frequencies of common words \citep{Argamon_2007,Burrows_2002}. 

Although stylometric approaches are useful for resolving certain types of authorship attribution tasks, there are clear limitations with these techniques. Overall performance declines dramatically when the number of candidate authors increases \citep{Grieve_2007,Luyckx_Daelemans_2011}, when the length of the question document decreases \citep{Eder_2015,grieve2017}, and when the amount of training data from the candidate authors decreases \citep{Luyckx_Daelemans_2011,grieve2017}.

Recent research in authorship analysis has begun to explore the use of modern Large Language Models (LLMs) to address these issues. Examples include universal authorial embeddings using Siamese BERT \citep{Rivera-Soto_Miano_Ordonez_Chen_Khan_Bishop_Andrews_2021} and Character BERT \cite{El_Boukkouri_Ferret_Lavergne_Noji_Zweigenbaum_Tsujii_2020}, and using BERT for classification \citep{Fabien_Villatoro-Tello_Motlicek_Parida_2020,Tyo_Dhingra_Lipton_2022}. LLM predictability metrics, such as perplexity and cross-entropy, have also been tested in a small number of studies. \citet{Fourkioti_Symeonidis_Arampatzis_2019} found that the perplexity of a single LLM pretrained on PoS-tags can be effective for authorship attribution. \citet{Barlas_Stamatatos_2020} extended this approach by training a multi-head classifier using the cross entropy of a single pretrained LLM, achieving their best performance using BERT, although they also considered other LLMs, including causal language models. Subsequently, \citet{Tyo_Dhingra_Lipton_2022} included \citet{Barlas_Stamatatos_2020}'s BERT-based approach, which they referred to as pALM (per Author Language Model), in their state-of-the-art authorship attribution benchmarking study, but found that pALM has the worst performance of all methods considered, which included an n-gram based classifier (Ngram) \citep{Tyo_Dhingra_Lipton_2022}, a prediction by partial matching compression model (PPM)\citep{croft_using_2003,neal_surveying_2018}, and a pre-trained BERT model with a dense layer for classification (BERT)\citep{Fabien_Villatoro-Tello_Motlicek_Parida_2020}. 

Although previous research has had relatively little success using LLM predictability metrics for human authorship attribution, this approach currently underlies state-of-the-art methods for LLM detection. LLM detection is a new type of attribution task that involves identifying whether a questioned text was written by a human or an LLM. The task has gained prominence in recent years due to increasing concerns about the misuse of LLMs \citep{Bommasani_Hudson_Adeli_Altman_Arora_von_Arx_Bernstein_Bohg_Bosselut_Brunskill__2022,Gehrmann_Strobelt_Rush_2019,Tian_Chen_Wang_Bai_Zhang_Li_Xu_Wang_2023,Wu_Pang_Shen_Cheng_Chua_2023,Gehrmann_Strobelt_Rush_2019,Wu_Pang_Shen_Cheng_Chua_2023}. In these studies, casual language model perplexity has been found to be an effective indicator of authorship, where LLM-authored texts tend to be associated with relatively low perplexity scores in comparison to human-authored texts: LLM-authored texts are generally more predictable to a LLM. Approaches include both fully automated detection and computer-assisted detection, e.g., GLTR \citep{Gehrmann_Strobelt_Rush_2019} and GPTZero \citep{chakraborty_possibilities_2023}. 

Building on this research, in this paper, we revisit the idea of using LLM perplexity for human authorship attribution. However, rather than work with a single LLM, as has been the case in previous research on both human and machine attribution, we build \textit{a set} of adapted \textit{Authorial Language Models} (ALMs), each of which is fine-tuned on the writings of a single candidate author. Instead of computing the perplexity or cross entropy based on a \textit{single} LLM, our approach involves predicting the authorship of a questioned document by comparing the perplexities of this text for \textit{multiple} ALMs, selecting the author whose associated LLM yields the lowest perplexity. We benchmark ALMs on the standard Blogs50 and CCAT50 datasets, following \citet{Tyo_Dhingra_Lipton_2022}, finding that our approach achieves state-of-the-art overall performance.

\section{Methodology}

\subsection{Authorship Attribution}

\textbf{Authorial LLM Fine-tuning} The first step of our approach for authorship attribution involves fine-tuning a set of causal language models based on the known writing of a set of candidate authors, creating one model for each author. 

For this study, we fine-tuned all authorial GPT-2 models with 100 epochs on a single Graphcore IPU Pod 4 Machine at PaperSpace. We make all scripts accessible online\footnote{\label{clmppl-repo-link}https://anonymous.4open.science/r/ALMs-4DC5}.

\textbf{Perplexity Calculation} The second step of our approach involves measuring the perplexity of a questioned text over each of the fine-tuned authorial language models to predict authorship.

The perplexity of a fixed-length causal language model \(M\) over a token sequence \(T=\left\{x_{1},x_{2},...,x_{t}\right\}\) is defined as
\[  ppl\left ( M,T \right ) = exp\left\{-\frac{1}{t}\sum_{i}^{t} log\left(p_{M}\left(  x_{i}|x_{< i}\right)\right) \right\}\]
In other words, perplexity is the exponentiated mean negative log likelihood of tokens in the sequence, which represents the average predictability of tokens in the sequence. 

In practice, we calculate perplexity as the cross entropy between the true token and the predicted logits, namely \(exp \left\{ CrossEntropy \left( Logits, T \right) \right\}\). Given a questioned text \(Q\), and a fine-tuned authorial GPT-2 model \(M\), we first pass \(Q\) to the GPT-2 BPE Tokenizer to extract a token sequence \(T\). \(T\) is then passed to \(M\) for language modeling, whose output is \(Logits\). Here \(Logits\) reflects the predicted probabilities of all tokens in \(T\), where \(T\) represents the ground truth. Therefore, in the next step, we measure the predictability of all tokens in \(T\) by comparing the predicted \(Logits\) and the ground truth \(T\) via cross entropy \(CE\), which we calculated using \(torch.nn.CrossEntropyLoss\) from PyTorch. Finally, we obtain the perplexity of \(Q\) under \(M\) as \(e\) raised to the power of \(CE\).

\textbf{Authorship Prediction} The third and final step of our approach involves attributing the questioned document to the author whose authorial LLM was associated with the lowest perplexity. Given a text \(Q\) from author \(i\), and a set of authorial language models \(\left\{M_{1},M_{2},...M_{n}\right\}\) fine-tuned on texts from a set of candidate authors \(\left\{1,2,..., n\right\}\), we expect \(ppl\left( M_{i}, Q \right)\) to be lowest among \(\left\{ppl\left( M_{1}, Q \right), ppl\left( M_{2}, Q \right), ..., ppl\left( M_{n}, Q \right)\right\}\), because we expect \(Q\) to be most predictable for the model that was fine-tuned on the training corpus for author \(i\).

\subsection{Data}

To evaluate ALMs, we used the Blogs50 \citep{Schler_Koppel_Argamon_Pennebaker_2006} and the CCAT50 \citep{Lewis_Yang_Rose_Li_2004} datasets. We chose these two datasets because they are accessible and allow us to benchmark our methods against the methods evaluated in \citet{Tyo_Dhingra_Lipton_2022}. Both CCAT50 and Blogs50 also contain sufficient numbers of authors, texts and tokens to adequately test and train our method and are focused on a relatively consistent registers: CCAT50 contains news articles, while Blogs50 contains texts from top bloggers. We retrieved both datasets from the repository of \citet{Tyo_Dhingra_Lipton_2022}\footnote{\label{valla-link}https://github.com/JacobTyo/Valla}. \autoref{dataset-facts} lists basic information for both datasets.

\begin{table}[h]
\centering
\begin{threeparttable}
\begin{tabular}{llllll}
\hline
Dataset&A&T&TK&T/A&TTL\\
\hline
 CCAT50 & 50&   5k&2.5M&100 &506\\
 Blogs50 & 50&   66k&8.1M&1.3k&122\\
\hline
\end{tabular}
\begin{tablenotes}
 \item A: author count; T: text count; TK: token count; 
 TTL: test text length, in token count
\end{tablenotes}
\end{threeparttable}
\caption{Information on Datasets}
\label{dataset-facts}
\end{table}

\section{Results}

To compare ALMs to state-of-the-art approaches for authorship attribution in NLP, we evaluated the performance of our method on Blogs50 dataset and CCAT50 dataset, using macro-average accuracy following \citet{Tyo_Dhingra_Lipton_2022}. \autoref{prev-aa-studies} compares the macro-average accuracy of our method against other methods. We find that our method outperforms all other methods on Blogs50, achieving a macro-average accuracy of 83.6\%, and outperforms all but one method on the CCAT 50 dataset, achieving a macro-average accuracy of 74.9\%, nearly matching the Ngram method, which achieves a macro-average accuracy of 76.7\%.

\begin{table}[h]
\centering
\begin{tabular}{lll}
\hline
Method& Blogs50&CCAT50 \\
\hline
ALMs& \textbf{83.6}\%&74.9\%\\
Ngram & 72.3\%&\textbf{76.7}\%\\
BERT & 75.0\%&65.7\%\\
PPM & 72.2\%& 69.4\%\\
pALM& -& 63.4\%\\
\hline
\end{tabular}
\caption{
Comparison Between Our Method (ALMs) And Recent SOTA Methods \citep{Tyo_Dhingra_Lipton_2022}}
\label{prev-aa-studies}
\end{table}

\begin{figure*}[!tbp]
    \centering
    \includegraphics[width=1\linewidth]{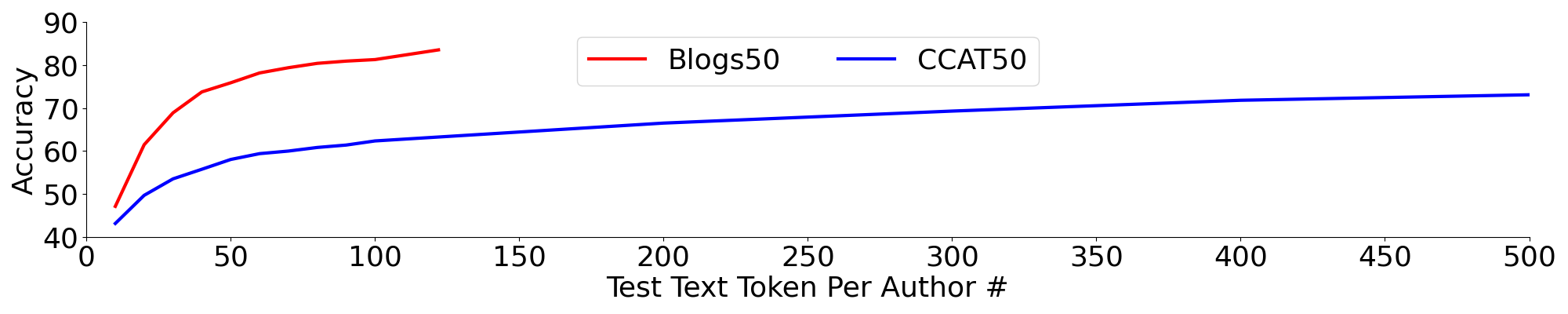}
    \caption{Performance of ALMs Under Different Test Text Lengths}
    \label{fig:ablation_test}
\end{figure*}


Furthermore, to evaluate the robustness of our method across individual authors, we calculated single author accuracy scores for the both the Blogs50 (see \autoref{author-by-author-score-blogs50}) and CCAT50 (see \autoref{author-by-author-score-ccat50}) datasets. For 38 out of the 50 authors in Blogs50 and for 31 out of the 50 authors in CCAT50 we obtained an accuracy of over 80\%. However, we note that there are a few authors whose texts prove especially difficult to attribute, including Author 46 in Blogs50 and Author 30 in CCAT50. 

In addition, we conducted an analysis of text ablation on the test texts from both datasets to assess the robustness of our method when working with questioned documents of a limited length. We find that to reach a macro-average accuracy of 70\%, ALMs requires test texts of at least 40 tokens on Blogs50 and of at least 400 tokens on CCAT50, while ALMs only needs a test text length of at least 20 tokens on Blogs50 and at least 70 tokens on CCAT50 to reach a macro-average accuracy of 60\%. We also plot the macro-average accuracy of ALMs under different test text lengths in \autoref{fig:ablation_test}, which shows not only that an increase in test text length results in higher macro-average accuracy scores, but that the performance of our approach falls off quickly on texts that contain fewer than 50 tokens.

\section{Discussion}

In this paper we proposed a new method for authorship attribution based on the perplexity of a set fine-tuned authorial causal language models that we refer to ALMs (Authorial Language Models). On the Blogs50 dataset our method achieves a macro-accuracy of 83.6\% which outperformed SOTA, while on the CCAT50 dataset our method achieves a macro-accuracy of 74.9\% nearly matching SOTA.

The performance of ALMs demonstrates that model predictability metrics from \textbf{multiple} authorial LLMs can be highly informative indicators in authorship attribution. Beyond fine-tuning for each author, using a large number of epochs and choosing causal language model of GPT-2 where perplexity is well defined may contribute to the excellent performance of ALMs compared to other techniques, including previous attempts to use LLM predictability metrics.

In addition, we attribute the excellent performance of ALMs compared to traditional stylometric methods based on its ability access to token-level authorial features. Compared to standard type-based methods in stylometry, which are based on the relative frequencies of common words, n-grams, and other types, perplexity-based methods are capable of capturing authorial information for each word token rather, thereby offering greater flexibility and finer granularity.

For instance, the use of the word \textit{baseball} is not generally a good feature for stylometric authorship attribution for two reasons. First, it is relatively infrequent, making it difficult to obtain a meaningful measurement of its relative frequency. This is why stylometric methods tend to focus on high frequency forms. Second, even if sufficient data were available, the relative frequency of this word type in any text would primarily reflect the topic of that text, as opposed to the style of its author. For example, a text with the frequent use of the word \textit{baseball} will tend to be about this sport. This is problematic in the context of authorship attribution because the goal is to attribute texts to the correct authors regardless of the topic. This is why stylometric methods tend to focus not only on common features, but on grammatical features, like function words. 

A token-based approach like ALMs, however, avoids these issues as it effectively assigns a probability to every token in a text. In general, we can assume that if a questioned document is about baseball, occurrences of the word \textit{baseball} will generally carry very little authorial information, and that the probability of the tokens of the word \textit{baseball} in that text will consistently be low for all authors. However, given a questioned document on some other topic, a token of the term \textit{baseball} (e.g., as an example or as a metaphor) would potentially be highly discriminatory – extremely unexpected for most authors, unless, for example, an author often uses baseball metaphors out of context.

In this sense, a token-based approach is similar to the type of qualitative stylistic authorship analysis often conducted manually in a forensic context, where forensic linguists examine a questioned document word by word \cite{coulthard2016introduction,grant2008approaching}. Like a forensic stylistic analysis, a great advantage of our approach compared to a standard stylometric analysis is that we can extract considerably more information from each text: every token is now a valid feature, whereas for traditional methods only frequent types can potentially be features. 

\section{Conclusion}

We developed an authorship attribution method ALMs that predicts author based on the perplexities of causal language models that fine-tuned on each of the candidate authors. ALMs achieves a macro-average accuracy score of 83.6\% on Blogs50, outperforming all other methods, and 74.9\% on CCAT50, matching the performance of the best method. Future research may focus on few-shot authorship attribution via perplexity of in-context-learning-capable LLM like Llama-2 \cite{touvron_llama_2023} and authorship profiling with filtered perplexity. In addition, we believe that our  perplexity-based approach constitutes a general method for comparative research in corpus linguistics, allowing for automated token-level comparative linguistic analysis. 

\section{Limitations}

Currently, our evaluation of ALMs was made under a fixed hyper-parameter settings where other combination of hyper-parameters remain untested. We also note that topical patterns in both datasets may positively influence the benchmarking scores for ALMs and other methods.

\section{Ethics and Impact}

Our research is based on publicly available base model; we are not aware of specific risks except biases inherited from data or base model, which needs to be examined before any implementations in a large scale. When put in practice, the predicted author from this method should be treated as reference to form the final decision of authorship together with other clues and evidences. In addition, it is important to stress that all methods for authorship analysis are limited by the quality of the dataset, and all methods for authorship attribution are limited by the completeness of the set of candidate authors. Finally, although we acknowledge that the computational cost of ALMs is relatively high, it is important to note that (1) stylometric approaches often out-perform human analysis especially when the amount of language data under analysis is relatively large, and (2) individual cases of disputed authorship can be extremely high value (e.g. in legal and security contexts). 

\section*{Acknowledgements}
This research is supported in part by the Office of the Director of National Intelligence (ODNI), Intelligence Advanced Research Projects Activity (IARPA), via the HIATUS Program contract \#2022-22072200006. The views and conclusions contained herein are those of the authors and should not be interpreted as necessarily representing the official policies, either expressed or implied, of ODNI, IARPA, or the U.S. Government. The U.S. Government is authorized to reproduce and distribute reprints for governmental purposes notwithstanding any copyright annotation therein.

\bibliography{mylib}

\begin{thebibliography}{28}
\expandafter\ifx\csname natexlab\endcsname\relax\def\natexlab#1{#1}\fi

\bibitem[{Argamon(2007)}]{Argamon_2007}
Shlomo Argamon. 2007.
\newblock \href {https://doi.org/10.1093/llc/fqn003} {Interpreting burrows’s delta: Geometric and probabilistic foundations}.
\newblock \emph{Literary and Linguistic Computing}, 23(2):131–147.

\bibitem[{Barlas and Stamatatos(2020)}]{Barlas_Stamatatos_2020}
Georgios Barlas and Efstathios Stamatatos. 2020.
\newblock \href {https://doi.org/10.1007/978-3-030-49161-1_22} {\emph{Cross-Domain Authorship Attribution Using Pre-trained Language Models}}, volume 583 of \emph{IFIP Advances in Information and Communication Technology}, page 255–266. Springer International Publishing, Cham.

\bibitem[{Binongo(2003)}]{Binongo_2003}
José Nilo~G. Binongo. 2003.
\newblock \href {https://doi.org/10.1080/09332480.2003.10554843} {Who wrote the 15th book of oz? an application of multivariate analysis to authorship attribution}.
\newblock \emph{Chance}, 16(2):9–17.

\bibitem[{Bommasani et~al.(2022)Bommasani, Hudson, Adeli, Altman, Arora, von Arx et~al.}]{Bommasani_Hudson_Adeli_Altman_Arora_von_Arx_Bernstein_Bohg_Bosselut_Brunskill__2022}
Rishi Bommasani, Drew~A. Hudson, Ehsan Adeli, Russ Altman, Simran Arora, Sydney von Arx, et~al. 2022.
\newblock \href {http://arxiv.org/abs/2108.07258} {On the opportunities and risks of foundation models}.
\newblock (arXiv:2108.07258).
\newblock ArXiv:2108.07258 [cs].

\bibitem[{Burrows(2002)}]{Burrows_2002}
John Burrows. 2002.
\newblock \href {https://doi.org/10.1093/llc/17.3.267} {“delta”: a measure of stylistic difference and a guide to likely authorship}.
\newblock \emph{Literary and Linguistic Computing}, 17(3):267–287.

\bibitem[{Chakraborty et~al.(2023)Chakraborty, Bedi, Zhu, An, Manocha, and Huang}]{chakraborty_possibilities_2023}
Souradip Chakraborty, Amrit~Singh Bedi, Sicheng Zhu, Bang An, Dinesh Manocha, and Furong Huang. 2023.
\newblock \href {http://arxiv.org/abs/2304.04736} {On the {Possibilities} of {AI}-{Generated} {Text} {Detection}}.
\newblock ArXiv:2304.04736 [cs].

\bibitem[{Coulthard et~al.(2016)Coulthard, Johnson, and Wright}]{coulthard2016introduction}
Malcolm Coulthard, Alison Johnson, and David Wright. 2016.
\newblock \emph{An introduction to forensic linguistics: Language in evidence}.
\newblock Routledge.

\bibitem[{Eder(2015)}]{Eder_2015}
Maciej Eder. 2015.
\newblock \href {https://doi.org/10.1093/llc/fqt066} {Does size matter? authorship attribution, small samples, big problem}.
\newblock \emph{Digital Scholarship in the Humanities}, 30(2):167–182.

\bibitem[{El~Boukkouri et~al.(2020)El~Boukkouri, Ferret, Lavergne, Noji, Zweigenbaum, and Tsujii}]{El_Boukkouri_Ferret_Lavergne_Noji_Zweigenbaum_Tsujii_2020}
Hicham El~Boukkouri, Olivier Ferret, Thomas Lavergne, Hiroshi Noji, Pierre Zweigenbaum, and Jun’ichi Tsujii. 2020.
\newblock \href {https://doi.org/10.18653/v1/2020.coling-main.609} {Characterbert: Reconciling elmo and bert for word-level open-vocabulary representations from characters}.
\newblock In \emph{Proceedings of the 28th International Conference on Computational Linguistics}, page 6903–6915, Barcelona, Spain (Online). International Committee on Computational Linguistics.

\bibitem[{Fabien et~al.(2020)Fabien, Villatoro-Tello, Motlicek, and Parida}]{Fabien_Villatoro-Tello_Motlicek_Parida_2020}
Maël Fabien, Esaú Villatoro-Tello, Petr Motlicek, and Shantipriya Parida. 2020.
\newblock Bertaa: Bert fine-tuning for authorship attribution.
\newblock In \emph{Proceedings of the 17th International Conference on Natural Language Processing (ICON)}, page 127–137.

\bibitem[{Fourkioti et~al.(2019)Fourkioti, Symeonidis, and Arampatzis}]{Fourkioti_Symeonidis_Arampatzis_2019}
Olga Fourkioti, Symeon Symeonidis, and Avi Arampatzis. 2019.
\newblock \href {https://doi.org/10.1016/j.ipm.2019.102061} {Language models and fusion for authorship attribution}.
\newblock \emph{Information Processing \& Management}, 56(6):102061.

\bibitem[{Gehrmann et~al.(2019)Gehrmann, Strobelt, and Rush}]{Gehrmann_Strobelt_Rush_2019}
Sebastian Gehrmann, Hendrik Strobelt, and Alexander Rush. 2019.
\newblock \href {https://doi.org/10.18653/v1/P19-3019} {Gltr: Statistical detection and visualization of generated text}.
\newblock In \emph{Proceedings of the 57th Annual Meeting of the Association for Computational Linguistics: System Demonstrations}, page 111–116, Florence, Italy. Association for Computational Linguistics.

\bibitem[{Grant(2008)}]{grant2008approaching}
Tim Grant. 2008.
\newblock Approaching questions in forensic authorship analysis.
\newblock \emph{Dimensions of forensic linguistics}, 5:215--229.

\bibitem[{Grieve(2007)}]{Grieve_2007}
Jack Grieve. 2007.
\newblock \href {https://doi.org/10.1093/llc/fqm020} {Quantitative authorship attribution: An evaluation of techniques}.
\newblock \emph{Literary and Linguistic Computing}, 22(3):251–270.

\bibitem[{Grieve(2023)}]{grieve2023_language}
Jack Grieve. 2023.
\newblock \href {https://doi.org/doi:10.1515/cllt-2022-0040} {Register variation explains stylometric authorship analysis}.
\newblock \emph{Corpus Linguistics and Linguistic Theory}, 19(1):47--77.

\bibitem[{Grieve et~al.(2018)Grieve, Clarke, Chiang, Gideon, Heini, Nini, and Waibel}]{grieve2017}
Jack Grieve, Isobelle Clarke, Emily Chiang, Hannah Gideon, Annina Heini, Andrea Nini, and Emily Waibel. 2018.
\newblock \href {https://doi.org/10.1093/llc/fqy042} {{Attributing the Bixby Letter using n-gram tracing}}.
\newblock \emph{Digital Scholarship in the Humanities}, 34(3):493--512.

\bibitem[{Juola(2006)}]{Juola_2006}
Patrick Juola. 2006.
\newblock \emph{Authorship attribution}.
\newblock Foundations and trends in information retrieval. Now Publ, Boston, Mass.

\bibitem[{Lewis et~al.(2004)Lewis, Yang, Rose, and Li}]{Lewis_Yang_Rose_Li_2004}
David~D Lewis, Yiming Yang, Tony~G Rose, and Fan Li. 2004.
\newblock Rcv1: A new benchmark collection for text categorization research.
\newblock \emph{Journal of machine learning research}, 5:361--397.

\bibitem[{Luyckx and Daelemans(2011)}]{Luyckx_Daelemans_2011}
Kim Luyckx and Walter Daelemans. 2011.
\newblock \href {https://doi.org/10.1093/llc/fqq013} {The effect of author set size and data size in authorship attribution}.
\newblock \emph{Literary and Linguistic Computing}, 26(1):35–55.

\bibitem[{Neal et~al.(2018)Neal, Sundararajan, Fatima, Yan, Xiang, and Woodard}]{neal_surveying_2018}
Tempestt Neal, Kalaivani Sundararajan, Aneez Fatima, Yiming Yan, Yingfei Xiang, and Damon Woodard. 2018.
\newblock \href {https://doi.org/10.1145/3132039} {Surveying {Stylometry} {Techniques} and {Applications}}.
\newblock \emph{ACM Computing Surveys}, 50(6):1--36.

\bibitem[{Rivera-Soto et~al.(2021)Rivera-Soto, Miano, Ordonez, Chen, Khan, Bishop, and Andrews}]{Rivera-Soto_Miano_Ordonez_Chen_Khan_Bishop_Andrews_2021}
Rafael~A. Rivera-Soto, Olivia~Elizabeth Miano, Juanita Ordonez, Barry~Y. Chen, Aleem Khan, Marcus Bishop, and Nicholas Andrews. 2021.
\newblock \href {https://doi.org/10.18653/v1/2021.emnlp-main.70} {Learning universal authorship representations}.
\newblock In \emph{Proceedings of the 2021 Conference on Empirical Methods in Natural Language Processing}, page 913–919, Online and Punta Cana, Dominican Republic. Association for Computational Linguistics.

\bibitem[{Schler et~al.(2006)Schler, Koppel, Argamon, and Pennebaker}]{Schler_Koppel_Argamon_Pennebaker_2006}
Jonathan Schler, Moshe Koppel, Shlomo Argamon, and James~W Pennebaker. 2006.
\newblock Effects of age and gender on blogging.
\newblock In \emph{AAAI spring symposium: Computational approaches to analyzing weblogs}, volume~6, page 199–205.

\bibitem[{Stamatatos(2009)}]{Stamatatos_2009}
Efstathios Stamatatos. 2009.
\newblock \href {https://doi.org/10.1002/asi.21001} {A survey of modern authorship attribution methods}.
\newblock \emph{Journal of the American Society for Information Science and Technology}, 60(3):538–556.

\bibitem[{Teahan and Harper(2003)}]{croft_using_2003}
William~J. Teahan and David~J. Harper. 2003.
\newblock \href {https://doi.org/10.1007/978-94-017-0171-6_7} {Using {Compression}-{Based} {Language} {Models} for {Text} {Categorization}}.
\newblock In W.~Bruce Croft and John Lafferty, editors, \emph{Language {Modeling} for {Information} {Retrieval}}, pages 141--165. Springer Netherlands, Dordrecht.

\bibitem[{Tian et~al.(2023)Tian, Chen, Wang, Bai, Zhang, Li, Xu, and Wang}]{Tian_Chen_Wang_Bai_Zhang_Li_Xu_Wang_2023}
Yuchuan Tian, Hanting Chen, Xutao Wang, Zheyuan Bai, Qinghua Zhang, Ruifeng Li, Chao Xu, and Yunhe Wang. 2023.
\newblock \href {http://arxiv.org/abs/2305.18149} {Multiscale positive-unlabeled detection of ai-generated texts}.
\newblock (arXiv:2305.18149).
\newblock ArXiv:2305.18149 [cs].

\bibitem[{Touvron et~al.(2023)Touvron, Martin, Stone, Albert, Almahairi, Babaei et~al.}]{touvron_llama_2023}
Hugo Touvron, Louis Martin, Kevin Stone, Peter Albert, Amjad Almahairi, Yasmine Babaei, et~al. 2023.
\newblock \href {https://arxiv.org/abs/2307.09288v2} {Llama 2: {Open} {Foundation} and {Fine}-{Tuned} {Chat} {Models}}.

\bibitem[{Tyo et~al.(2022)Tyo, Dhingra, and Lipton}]{Tyo_Dhingra_Lipton_2022}
Jacob Tyo, Bhuwan Dhingra, and Zachary~C. Lipton. 2022.
\newblock \href {http://arxiv.org/abs/2209.06869} {On the state of the art in authorship attribution and authorship verification}.
\newblock (arXiv:2209.06869).
\newblock ArXiv:2209.06869 [cs].

\bibitem[{Wu et~al.(2023)Wu, Pang, Shen, Cheng, and Chua}]{Wu_Pang_Shen_Cheng_Chua_2023}
Kangxi Wu, Liang Pang, Huawei Shen, Xueqi Cheng, and Tat-Seng Chua. 2023.
\newblock \href {http://arxiv.org/abs/2305.15004} {Llmdet: A large language models detection tool}.
\newblock (arXiv:2305.15004).
\newblock ArXiv:2305.15004 [cs].

\end{thebibliography}
\appendix
\begin{table*}
\centering
\begin{tabular}{lllll}
\hline
Author Name & Text \# & Token \# & Mean Token \# per Text& Accuracy(\%)\\
\hline
0&	2623&	306575&	116.9&	90.85\\
1&	1405&	341521&	243.1&	94.89\\
2&	1311&	298754&	227.9&	86.89\\
3&	1301&	207581&	159.6&	83.69\\
4&	1215&	254051&	209.1&	96.05\\
5&	1207&	194987&	161.5&	92.38\\
6&	1125&	144410&	128.4&	86.48\\
7&	1100&	119820&	108.9&	95.64\\
8&	1083&	257212&	237.5&	83.39\\
9&	1078&	182824&	169.6&	89.63\\
10&	1046&	456278&	436.2&	88.17\\
11&	1019&	166745&	163.6&	43.53\\
12&	1009&	164719&	163.2&	73.41\\
13&	947&	132354&	139.8&	88.19\\
14&	910&	191351&	210.3&	91.67\\
15&	894&	211436&	236.5&	95.09\\
16&	835&	237192&	284.1&	80.86\\
17&	830&	162396&	195.7&	84.13\\
18&	811&	288834&	356.1&	95.07\\
19&	808&	63683&	78.8&	99.5\\
20&	807&	160624&	199&	91.58\\
21&	795&	132298&	166.4&	84.92\\
22&	782&	290040&	370.9&	90.77\\
23&	755&	331710&	439.4&	83.07\\
24&	753&	91592&	121.6&	83.51\\
25&	743&	213729&	287.7&	76.34\\
26&	740&	102774&	138.9&	67.57\\
27&	740&	148279&	200.4&	98.92\\
28&	726&	111685&	153.8&	83.43\\
29&	726&	96883&	133.4&	75.14\\
30&	718&	87430&	121.8&	88.27\\
31&	716&	499373&	697.4&	97.77\\
32&	707&	128940&	182.4&	76.27\\
33&	706&	181419&	257&	88.64\\
34&	705&	118039&	167.4&	80.68\\
35&	689&	79639&	115.6&	83.72\\
36&	661&	87402&	132.2&	64.24\\
37&	636&	203631&	320.2&	98.74\\
38&	631&	359944&	570.4&	93.04\\
39&	621&	248852&	400.7&	67.1\\
40&	609&	50015&	82.1&	82.24\\
41&	609&	128077&	210.3&	88.16\\
42&	605&	81499&	134.7&	75.5\\
43&	605&	90913&	150.3&	84.77\\
44&	605&	456833&	755.1&	96.03\\
45&	600&	78737&	131.2&	50\\
46&	593&	86792&	146.4&	39.19\\
47&	592&	132766&	224.3&	60.81\\
48&	576&	202813&	352.1&	94.44\\
49&	565&	118027&	208.9&	95.04\\
\hline
\end{tabular}
\caption{Performance of Our Method for Each of the 50 Authors in Blogs50}
\label{author-by-author-score-blogs50}
\end{table*}

\begin{table*}
\centering
\begin{tabular}{lllll}
\hline
Author Name & Text \# & Token \# & Mean Token \# per Text& Accuracy(\%)\\
\hline
0&	90&	55807&	620.1&	50.00\\
1&	90&	63795&	708.8&	68.00\\
2&	90&	65540&	728.2&	66.00\\
3&	90&	62633&	695.9&	88.00\\
4&	90&	63459&	705.1&	96.00\\
5&	90&	62785&	697.6&	88.00\\
6&	90&	58634&	651.5&	90.00\\
7&	90&	51316&	570.2&	82.00\\
8&	90&	59730&	663.7&	30.00\\
9&	90&	57452&	638.4&	88.00\\
10&	90&	57416&	638&	92.00\\
11&	90&	54029&	600.3&	40.00\\
12&	90&	61247&	680.5&	60.00\\
13&	90&	58148&	646.1&	34.00\\
14&	90&	53918&	599.1&	80.00\\
15&	90&	53871&	598.6&	82.00\\
16&	90&	65597&	728.9&	60.00\\
17&	90&	65092&	723.2&	30.00\\
18&	90&	53021&	589.1&	84.00\\
19&	90&	70496&	783.3&	80.00\\
20&	90&	57850&	642.8&	78.00\\
21&	90&	64731&	719.2&	94.00\\
22&	90&	68547&	761.6&	80.00\\
23&	90&	61034&	678.2&	88.00\\
24&	90&	69984&	777.6&	100.00\\
25&	90&	56861&	631.8&	92.00\\
26&	90&	55312&	614.6&	90.00\\
27&	90&	66168&	735.2&	68.00\\
28&	90&	52956&	588.4&	86.00\\
29&	90&	57617&	640.2&	86.00\\
30&	90&	65783&	730.9&	14.00\\
31&	90&	43818&	486.9&	82.00\\
32&	90&	53929&	599.2&	90.00\\
33&	90&	59234&	658.2&	60.00\\
34&	90&	66891&	743.2&	28.00\\
35&	90&	63705&	707.8&	100.00\\
36&	90&	59393&	659.9&	80.00\\
37&	90&	64213&	713.5&	78.00\\
38&	90&	53000&	588.9&	62.00\\
39&	90&	58239&	647.1&	48.00\\
40&	90&	54635&	607.1&	86.00\\
41&	90&	68160&	757.3&	82.00\\
42&	90&	56512&	627.9&	98.00\\
43&	90&	63766&	708.5&	92.00\\
44&	90&	62480&	694.2&	42.00\\
45&	90&	56889&	632.1&	92.00\\
46&	90&	57731&	641.5&	72.00\\
47&	90&	55562&	617.4&	100.00\\
48&	90&	67006&	744.5&	88.00\\
49&	90&	41518&	461.3&	100.00\\
\hline
\end{tabular}
\caption{Performance of Our Method for Each of the 50 Authors in CCAT50}
\label{author-by-author-score-ccat50}
\end{table*}

\end{document}